\documentclass[conference]{IEEEtran}
\IEEEoverridecommandlockouts
\usepackage{cite}
\usepackage{amsmath,amssymb,amsfonts}
\usepackage{algorithmic}
\usepackage{graphicx}
\usepackage{textcomp}
\usepackage{xcolor}
\def\BibTeX{{\rm B\kern-.05em{\sc i\kern-.025em b}\kern-.08em
    T\kern-.1667em\lower.7ex\hbox{E}\kern-.125emX}}

\usepackage{hyperref}
\hypersetup{colorlinks,linkcolor={blue},citecolor={blue},urlcolor={red}}  

\begin{document}

\title{An Evaluation of Forensic Facial Recognition \\
\thanks{We are grateful to the team at Synthesis AI (\url{https://synthesis.ai}) for generously providing access to their image synthesis API.}}

\author{\IEEEauthorblockN{Justin Norman}
\IEEEauthorblockA{\textit{School of Information} \\
\textit{University of California, Berkeley}\\
Berkeley, CA, USA \\
justin.norman@berkeley.edu}
\and
\IEEEauthorblockN{Shruti Agarwal}
\IEEEauthorblockA{\textit{Adobe Inc.} \\
San Jose, CA, USA \\
shragarw@adobe.com}
\and
\IEEEauthorblockN{Hany Farid}
\IEEEauthorblockA{\textit{EECS and School of Information} \\
\textit{University of California, Berkeley}\\
Berkeley, CA, USA \\
hfarid@berkeley.edu}
}

\maketitle

\begin{abstract}
Recent advances in machine learning and computer vision have led to reported facial recognition accuracies surpassing human performance. We question if these systems will translate to real-world forensic scenarios in which a potentially low-resolution, low-quality, partially-occluded image is compared against a standard facial database. We describe the construction of a large-scale synthetic facial dataset along with a controlled facial forensic lineup, the combination of which allows for a controlled evaluation of facial recognition under a range of real-world conditions. Using this synthetic dataset, and a popular dataset of real faces, we evaluate the accuracy of two popular neural-based recognition systems. We find that previously reported face recognition accuracies of more than $95\%$ drop to as low as $65\%$ in this more challenging forensic scenario.
\end{abstract}


%-------------------------------------------------------------------------
\section{Introduction}
\label{sec:introduction}

Some of the earliest approaches to automatic facial recognition date back to the mid 1960s~\cite{bledsoe1966model,bledsoe1966man}. It wasn't until some three decades later, however, that automatic facial recognition entered the mainstream with the publication of Turk and Pentland's seminal Eigenfaces~\cite{turk1991eigenfaces}. Over the next three decades, facial recognition moved from a largely academic pursuit to the mainstream. Today, facial recognition is widely used on mobile devices, social-media platforms, department store surveillance, border crossings, and by law enforcement and governments around the world.

The commercial (and controversial) Clearview AI, for example, has scraped billions of online photos to create a global facial database. Clearview's recognition software then allows their clients (which include government and law-enforcement agencies) to upload a photo to be matched against their massive database. 

Clearview has heralded this technology as a boon for public safety, while others have bemoaned its contribution to an increasing surveillance state~\cite{roussi2020resisting}. Clearview highlights a case where US federal agents arrested a man suspected of the sexual abuse of a seven-year old girl by matching a suspect's photo against a person reflected in a gym mirror in the background of an unrelated photo~\cite{taboh2021clearview}. This facial match eventually led to the identification of the gym's location and the suspect's arrest and conviction.

On the other hand, troubling reports of the use of facial recognition to fuel human-rights violations have also emerged~\cite{bbc22-thefaces}, as have concerns over privacy for those who have not consented to have their faces added to a global database~\cite{hill2020secretive,smith2022ethical}. 

Much has been written about flaws in facial recognition, particularly in terms of gender and racial bias~\cite{buolamwini2018gender,raji2020saving,chouldechova2022unsupervised}. With facial recognition systems seeing widespread use in law enforcement, it is also critical that we understand its accuracy, particularly in high-stakes forensic settings.

Consider a facial recognition system that seeks to identify person $X$ in a database with $N$ instances of $X$. The system returns $n$ correct instances of $X$ (true positive) and $m$ incorrect instances of $X$ (false positive). The precision of this system is $n/(n+m)$; and the recall is $n/N$. How we assess the importance of these two measures of accuracy depends on how the system is being used. In a preliminary investigatory setting, we may tolerate a lower precision in favor of a higher recall, allowing a downstream investigation to confirm any identifications with additional evidence. On the other hand, in a criminal evidentiary setting where the government is presenting incriminating photographic evidence of a crime, we should demand a high precision to ensure an innocent person is not falsely accused. 

\begin{figure*}[t!]
    \begin{center}
    \begin{tabular}{c@{\hspace{0.1cm}}|@{\hspace{0.1cm}}c@{\hspace{0.1cm}}c@{\hspace{0.1cm}}c@{\hspace{0.1cm}}c@{\hspace{0.1cm}}c@{\hspace{0.1cm}}c}
    probe & lineup$^*$ & lineup & lineup & lineup & lineup & lineup \\
    \includegraphics[width=0.125\textwidth]{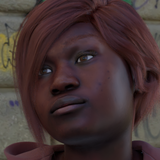} &
    \includegraphics[width=0.125\textwidth]{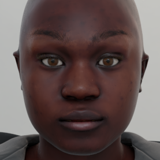} &
    \includegraphics[width=0.125\textwidth]{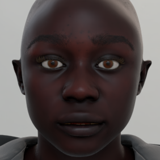} &
    \includegraphics[width=0.125\textwidth]{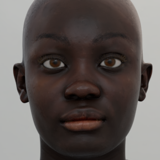} &
    \includegraphics[width=0.125\textwidth]{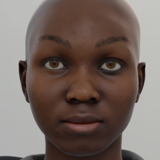} &
    \includegraphics[width=0.125\textwidth]{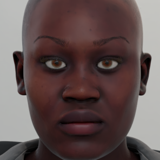} &
    \includegraphics[width=0.125\textwidth]{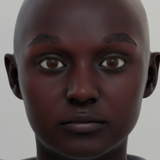} \\
    \includegraphics[width=0.125\textwidth]{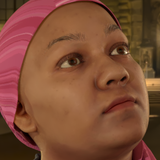} &
    \includegraphics[width=0.125\textwidth]{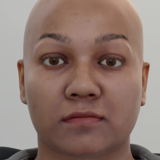} &
    \includegraphics[width=0.125\textwidth]{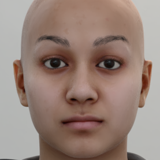} &
    \includegraphics[width=0.125\textwidth]{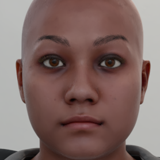} &
    \includegraphics[width=0.125\textwidth]{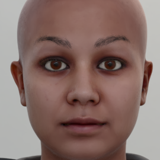} &
    \includegraphics[width=0.125\textwidth]{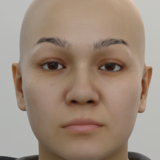} &
    \includegraphics[width=0.125\textwidth]{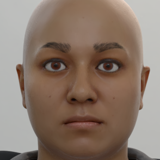} \\
    \includegraphics[width=0.125\textwidth]{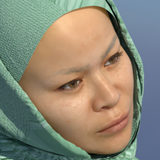} &
    \includegraphics[width=0.125\textwidth]{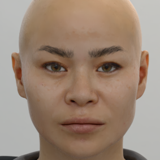} &
    \includegraphics[width=0.125\textwidth]{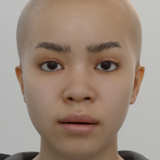} &
    \includegraphics[width=0.125\textwidth]{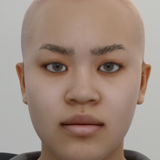} &
    \includegraphics[width=0.125\textwidth]{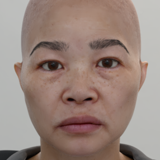} &
    \includegraphics[width=0.125\textwidth]{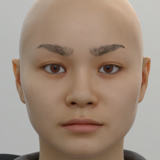} &
    \includegraphics[width=0.125\textwidth]{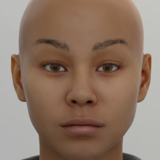} \\
    \includegraphics[width=0.125\textwidth]{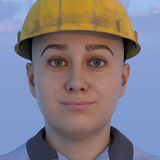} &
    \includegraphics[width=0.125\textwidth]{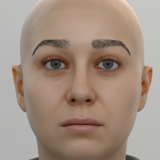} &
    \includegraphics[width=0.125\textwidth]{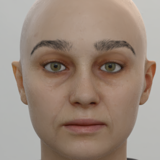} &
    \includegraphics[width=0.125\textwidth]{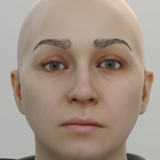} &
    \includegraphics[width=0.125\textwidth]{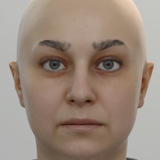} &
    \includegraphics[width=0.125\textwidth]{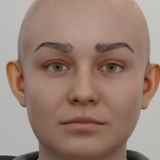} &
    \includegraphics[width=0.125\textwidth]{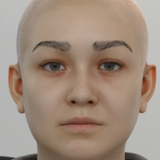} \\
    \includegraphics[width=0.125\textwidth]{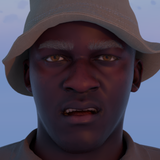} &
    \includegraphics[width=0.125\textwidth]{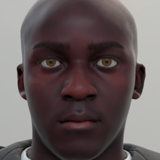} &
    \includegraphics[width=0.125\textwidth]{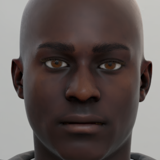} &
    \includegraphics[width=0.125\textwidth]{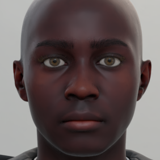} &
    \includegraphics[width=0.125\textwidth]{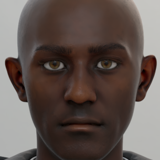} &
    \includegraphics[width=0.125\textwidth]{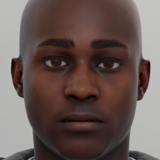} &
    \includegraphics[width=0.125\textwidth]{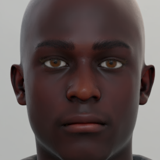} \\
    \includegraphics[width=0.125\textwidth]{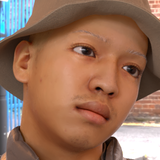} &
    \includegraphics[width=0.125\textwidth]{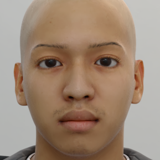} &
    \includegraphics[width=0.125\textwidth]{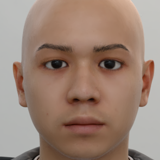} &
    \includegraphics[width=0.125\textwidth]{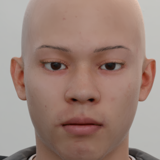} &
    \includegraphics[width=0.125\textwidth]{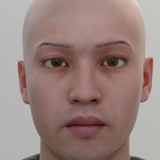} &
    \includegraphics[width=0.125\textwidth]{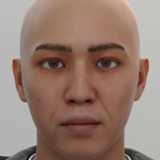} &
    \includegraphics[width=0.125\textwidth]{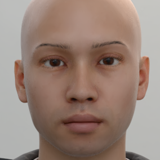} \\
    \includegraphics[width=0.125\textwidth]{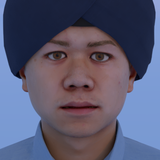} &
    \includegraphics[width=0.125\textwidth]{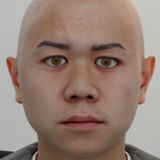} &
    \includegraphics[width=0.125\textwidth]{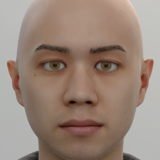} &
    \includegraphics[width=0.125\textwidth]{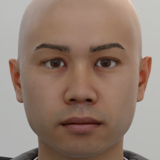} &
    \includegraphics[width=0.125\textwidth]{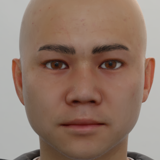} &
    \includegraphics[width=0.125\textwidth]{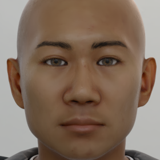} &
    \includegraphics[width=0.125\textwidth]{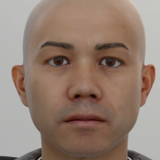} \\
    \includegraphics[width=0.125\textwidth]{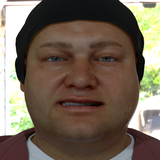} &
    \includegraphics[width=0.125\textwidth]{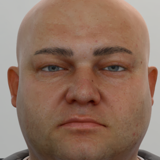} &
    \includegraphics[width=0.125\textwidth]{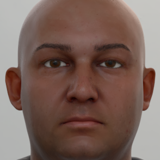} &
    \includegraphics[width=0.125\textwidth]{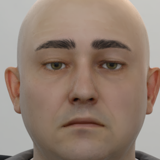} &
    \includegraphics[width=0.125\textwidth]{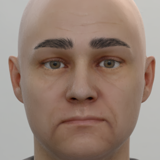} &
    \includegraphics[width=0.125\textwidth]{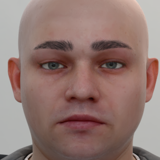} &
    \includegraphics[width=0.125\textwidth]{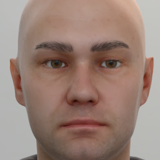} \\
    \end{tabular}
    \end{center}
    \vspace{-0.3cm}
    \caption{Examples of facial forensic lineups consisting of a probe image (left) and six standardized images, one of which (*) matches the identity in the probe image, and the rest of which are decoys. These images are from our synthetically-generated data set consisting of $200,000$ images and $8,000$ identities.}
    \label{fig:synthetic-faces}
\end{figure*}

While precision and recall can be a reasonable assessment of overall accuracy, an often overlooked aspect of these measurements is the composition of the comparison group. In the above example a high precision may be relatively easy to achieve if the person $X$ has highly distinct characteristics (age, race, gender, etc.) as compared to the database against which they are being compared. On the other hand, the same system may struggle if person $X$ shares many characteristics with the comparison group. Also often overlooked is the prevalence of each identity in the reference database: if multiple examples of $X$ are present in the database, then a match is much more likely.

We propose that these variations can be better controlled for by evaluating facial recognition within a well-defined lineup. In particular, in the classic eyewitness setting, a witness is asked to identify a suspect in a six-person lineup consisting of the suspect and five decoys with the same general characteristics and distinguishing features as the suspect~\cite{wixted2021eyewitness}. We describe a similar approach to assess the accuracy of a facial recognition system deployed in a forensic setting. 

We first describe the construction of a large-scale, synthetic dataset of $200,000$ images consisting of $8,000$ distinct identities, spanning different head poses and accessories. Shown in Fig.~\ref{fig:synthetic-faces} are representative examples of these images. We will show that these synthetic identities are a reasonable facsimile to real identities by comparing facial recognition accuracy between this dataset and the CASIA-Webface dataset of $494,414$ images consisting of $10,575$ identities. 

An added benefit of this large-scale synthetic dataset is that it allows us to evaluate facial recognition accuracy under controlled and varied conditions including scene variations (head pose, head wear, sunglasses, and masks) and image degradations  (blur, resolution, noise, and compression quality).

We evaluate two popular systems (ArcFace~\cite{deng2019arcface} and FaceNet~\cite{schroff2015facenet}) on a facial lineup recognition task under varying degrees of similarity in lineup identities, variations in head pose and occluding glasses and masks, and a range of image degradations.

Although much has been written about the strengths and limitations of automatic facial recognition (e.g.,~\cite{o2021face,wang2021deep}), to our knowledge, our analysis is the first that specifically addresses the unique requirements of {\em forensic} facial recognition in which by carefully controlling the composition of the comparison group, we provide a more realistic assessment of how (or even if) facial recognition systems should be used by law enforcement.

% FYI: https://www.law.georgetown.edu/privacy-technology-center/publications/a-forensic-without-the-science-face-recognition-in-u-s-criminal-investigations/

%-------------------------------------------------------------------------
\section{Face Recognition}
\label{sec:face-recognition}

The accuracy and robustness of facial recognition have improved drastically over the last few years, mainly due to powerful neural-network architectures~\cite{Parkhi15}, robust loss functions~\cite{deng2019arcface}, and large-scale datasets~\cite{bae2022digiface}. We make use of two popular facial recognition systems, FaceNet~\cite{schroff2015facenet} and ArcFace~\cite{deng2019arcface}. Although only these two architectures are evaluated, our general approach can be extended to evaluate any facial recognition system. 

FaceNet~\cite{schroff2015facenet} uses an inception ResnetV1-based model architecture, trained and evaluated on either the CASIA-WebFace or VGGFace2 datasets. The network yields a $128$D embedding from each input image. This results in an output such that the squared L2 distances in embedding space represent face similarity, where similar faces have small distances and dissimilar faces have large distances. The original FaceNet architecture was tested on the LFW dataset~\cite{huang2008labeled} and the Sandberg implementation~\cite{Sandberg2018}.

ArcFace~\cite{deng2019arcface} utilizes a $512$-D normalized embedding feature, organized into distinct cluster representing individual identities. The model architecture then employs an additive angular margin loss yielding better identity separability and, in turn, recognition accuracy. ArcFace is trained and evaluated on CASIA-Webface, Visual Geometry Group Face Dataset 2 (VGGFace2)\cite{Cao18} and a curated and tightly-cropped-to-faces version of MS1MV0~\cite{bae2022digiface} for comparison purposes.

%-------------------------------------------------------------------------
\section{Facial Datasets}
\label{sec:facial-datasets}

Most of the state-of-the-art facial recognition systems, including FaceNet and ArcFace, are trained on large-scale image datasets collected from a variety of online sources. These in-the-wild datasets, however, do not provide fine-grained control over differences in appearance within and across identities.

Recently, it has been shown that synthetically-generated faces can be used to successfully train a facial recognition system that then generalizes to real-world faces. These synthetic images can be generated using either a classic graphics pipeline~\cite{bae2022digiface} or a generative network~\cite{qiu2021synface}.

The graphics pipeline provides better control over a large number of facial attributes, and does not give rise to privacy concerns that emerge from a generative approach which relies on images of real people~\cite{bae2022digiface}. We employ Synthesis AI's commercial software (\url{https://synthesis.ai}) which uses a combination of classic rendering and generative synthesis to create photorealistic human faces. All images were rendered at a resolution of $512 \times 512$ pixels (and eventually downsized to $160 \times 160$ as input to FaceNet and ArcFace). A total of $200,000$ images were rendered consisting of $8,000$ unique identities with varying head poses, expressions, head wear, facial hair, hairstyles, glasses (opaque and clear), masks, backgrounds, and environmental lighting, Fig.~\ref{fig:synthetic-faces}.

Unlike previous approaches in which synthetic data is used for {\em training} a facial-recognition system, we use synthetic data to {\em evaluate} the reliability of these systems. 

The real-world dataset is derived from the CASIA-Webface dataset, consisting of $453,453$ images derived from $10,575$ identities. These images are of various size, quality, pose, subject clothing, and environment. Due to the initial quality of the dataset, some manual curation was necessary including the removal of duplicates, the removal of incorrectly labeled images, and standardization of image size to a resolution of $160 \times 160$ pixels.

%-------------------------------------------------------------------------
\section{Forensic Face Recognition}
\label{sec:forensic-face-recognition}

We emulate a real-world lineup where an image of a suspect--the probe--is placed in a six-person lineup consisting of five decoys with the same general characteristics as the suspect along with an image of the suspect in a different pose, expression, lighting, etc., Fig.~\ref{fig:synthetic-faces}. In order to capture typical real-world situations, we configure the lineup so that the probe image is unconstrained (as might be recorded from CCTV), but the lineup images are all standard mug-shot style photos (frontal view, no glasses or mask, and uniform lighting).

The lineup decoys are selected by searching a dataset of faces (Section~\ref{sec:facial-datasets}) for perceptually similar faces. Using the multi-task cascaded convolutional network architecture~\cite{zhang2016MTCNN} as implemented by Sandberg~\cite{Sandberg2018}, a tight square bounding box is placed around the face in each image, after which each image is cropped to this bounding box, and then resized to $160 \times 160$ pixels. A latent representation is extracted from each cropped face using the corresponding (FaceNet or ArcFace) architecture~\cite{schroff2015facenet}. The perceptual similarity of two faces is then computed as the cosine similarity between a pair of latent representations. A digital lineup for a probe identity $X$ is created by selecting the five distinct identities that are most similar to $X$, and a different image of $X$, Fig.~\ref{fig:synthetic-faces}. The facial match is correct if the cosine similarity between identity $X$ in the probe and lineup is more than than all five decoys.

To emulate real-world conditions, we will consider a range of scene and image degradations to the probe image and their impact on facial recognition. We will also explore any potential racial or gender bias.

%-------------------------------------------------------------------------
% https://docs.google.com/spreadsheets/d/1z9L8IV7uIyDjJYjcfD_o6HfiXNN-_IZy8wTbblxU3mk/edit#gid=0
%

%
%
\begin{figure*}[t]
    \begin{center}
    \begin{tabular}{rc@{\hspace{0.1cm}}c@{\hspace{0.1cm}}c@{\hspace{0.1cm}}c@{\hspace{0.1cm}}c}
     & 1 & 5 & 9 & 13 & 17 \\
     \raisebox{1.4cm}{blur} & 
     \includegraphics[width=0.15\linewidth]{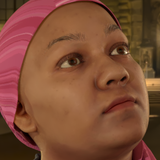} &
     \includegraphics[width=0.15\linewidth]{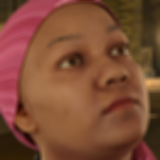} &
     \includegraphics[width=0.15\linewidth]{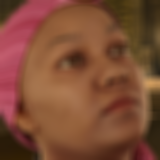} &
     \includegraphics[width=0.15\linewidth]{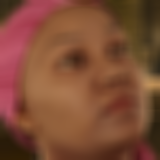} &
     \includegraphics[width=0.15\linewidth]{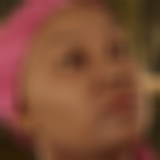} \\
     & 0.8625 & 0.6625 & 0.4625 & 0.2625 & 0.0625 \\
     \raisebox{1.4cm}{scale} & 
     \includegraphics[width=0.129\linewidth]{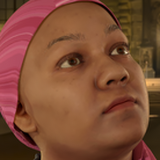} &
     \includegraphics[width=0.099\linewidth]{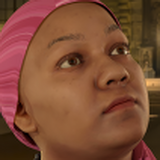} &
     \includegraphics[width=0.069\linewidth]{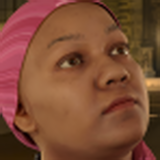} &
     \includegraphics[width=0.039\linewidth]{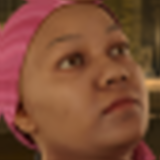} &
     \includegraphics[width=0.009\linewidth]{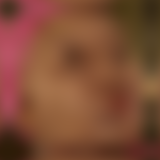} \\
     & 16 & 8 & 0 & -16 & -8 \\
     \raisebox{1.4cm}{noise} & 
     \includegraphics[width=0.15\linewidth]{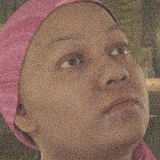} &
     \includegraphics[width=0.15\linewidth]{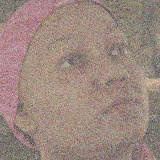} &
     \includegraphics[width=0.15\linewidth]{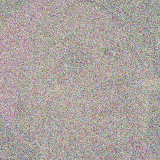} &
     \includegraphics[width=0.15\linewidth]{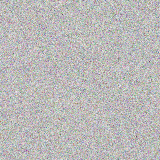} &
     \includegraphics[width=0.15\linewidth]{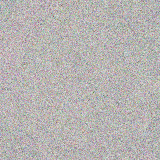} \\
     & 0.9 & 0.7 & 0.5 & 0.3 & 0.1 \\
    \raisebox{1.4cm}{jpeg} & 
    \includegraphics[width=0.15\linewidth]{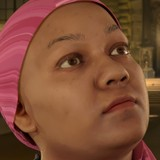} &
    \includegraphics[width=0.15\linewidth]{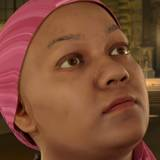} &
    \includegraphics[width=0.15\linewidth]{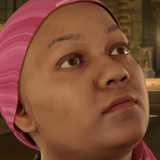} &
    \includegraphics[width=0.15\linewidth]{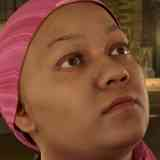} &
    \includegraphics[width=0.15\linewidth]{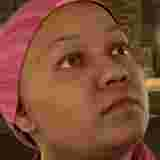} \\
    & 0.05 & 0.3 & 1.3 & 3.3 & 5.3 \\
    \raisebox{1.4cm}{gamma} &
    \includegraphics[width=0.15\linewidth]{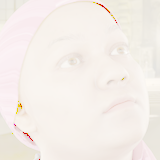} &
    \includegraphics[width=0.15\linewidth]{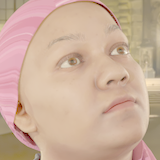} &
    \includegraphics[width=0.15\linewidth]{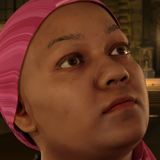} &
    \includegraphics[width=0.15\linewidth]{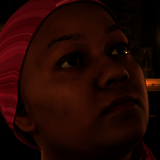} &
    \includegraphics[width=0.15\linewidth]{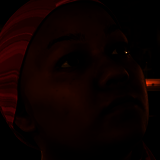} 
    \end{tabular}
    \end{center}
    \vspace{-0.3cm}
\caption{Examples of image degradation applied to a synthetically-generated face. See Fig.~\ref{fig:results-facenet-arcface}.}
\label{fig:degradations}
\end{figure*}

\section{Evaluation}
\label{sec:evaluation}

On the real-world CASIA-Webface facial dataset (Section~\ref{sec:facial-datasets}), facial recognition accuracy on our forensic lineup task is $73.1\%$ for FaceNet and $82.7\%$ for ArcFace (chance performance on a lineup with six identities is $1/6=16.7\%$). This accuracy on a forensic lineup stands in contrast to previously reported accuracies of between $95\%$ and $99\%$ for FaceNet~\cite{schroff2015facenet} and ArcFace~\cite{deng2019arcface}. 

We hypothesize that the difference in these accuracies is due to our more challenging forensic lineup task in which the probe image must be matched to exactly one other image in a lineup with the same identity. In comparison, comparing a probe image to the entire dataset--potentially containing multiple instances of the same identity--is an easier task because success is accomplished by matching against any of $N$ instances of the same identity.

For our synthetic facial dataset (Section~\ref{sec:facial-datasets}), facial recognition accuracy is $75.5\%$ for FaceNet, on par with the real-world dataset; and $95.6\%$ for ArcFace, significantly better than the real-world dataset. We hypothesize that this synthetic/real difference is the result of the fact that ArcFace explicitly removes noisy images from their training and therefore their model is optimized for relatively noise-free images. Despite these synthetic/real differences, we will show that by manipulating the quality of the probe image used to create the lineup, we can realign the accuracy on the synthetic and real-world datasets.

Although not an issue here, if the accuracy on the synthetic dataset was lower than the real-world accuracy, the lineup selection process could be modified to select the $n^{th}$--$(n-4)^{th}$ most similar faces for the lineup, instead of the top-five most similar faces, thus making the forensic lineup task easier.

\subsection{Image Degradations}

\begin{figure}[t!]
    \begin{tabular}{c@{\hspace{0.1cm}}c}
    FaceNet & ArcFace \\
    \includegraphics[width=4.4cm]{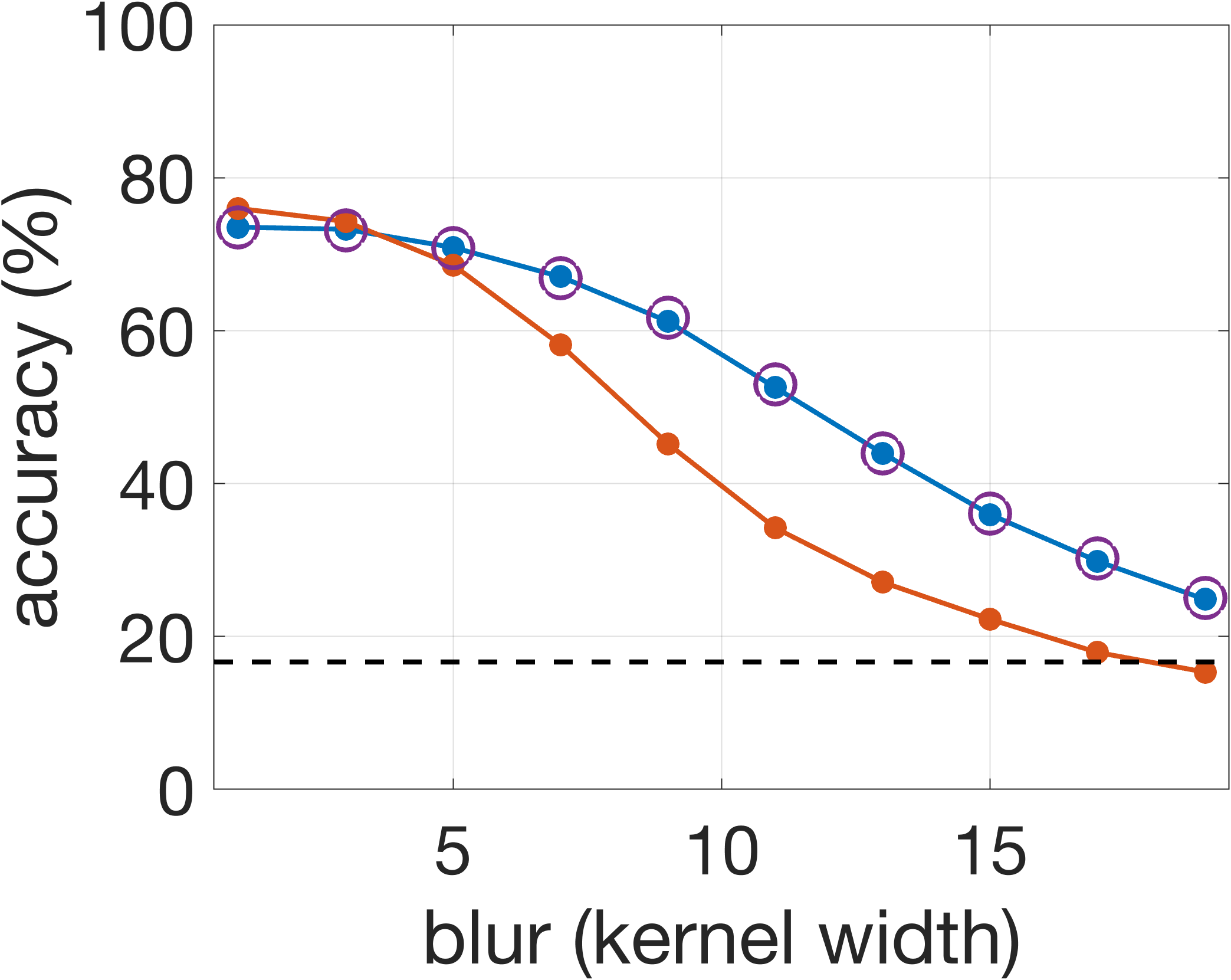}  & 
    \includegraphics[width=3.82cm]{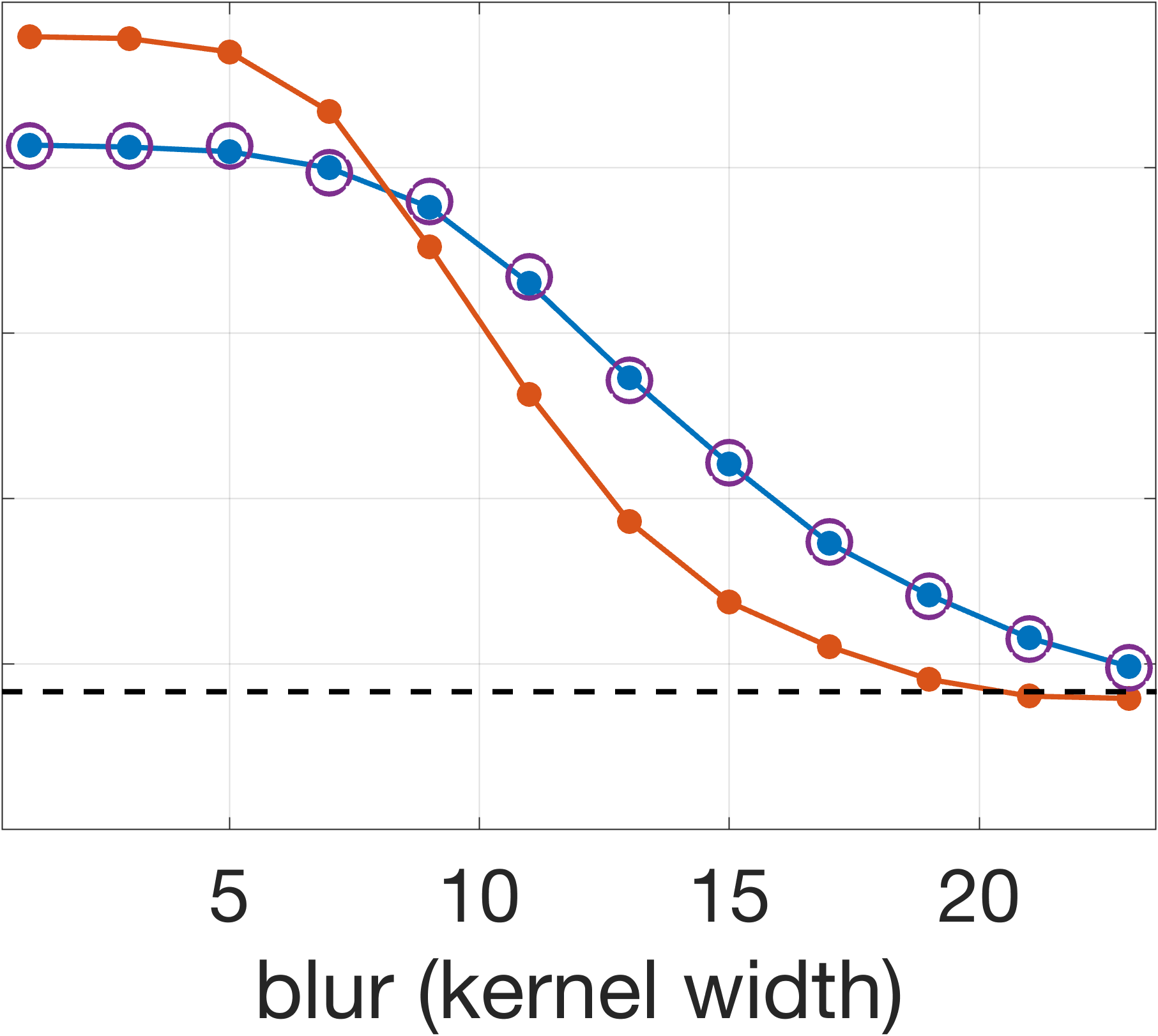}  \\
    \includegraphics[width=4.4cm]{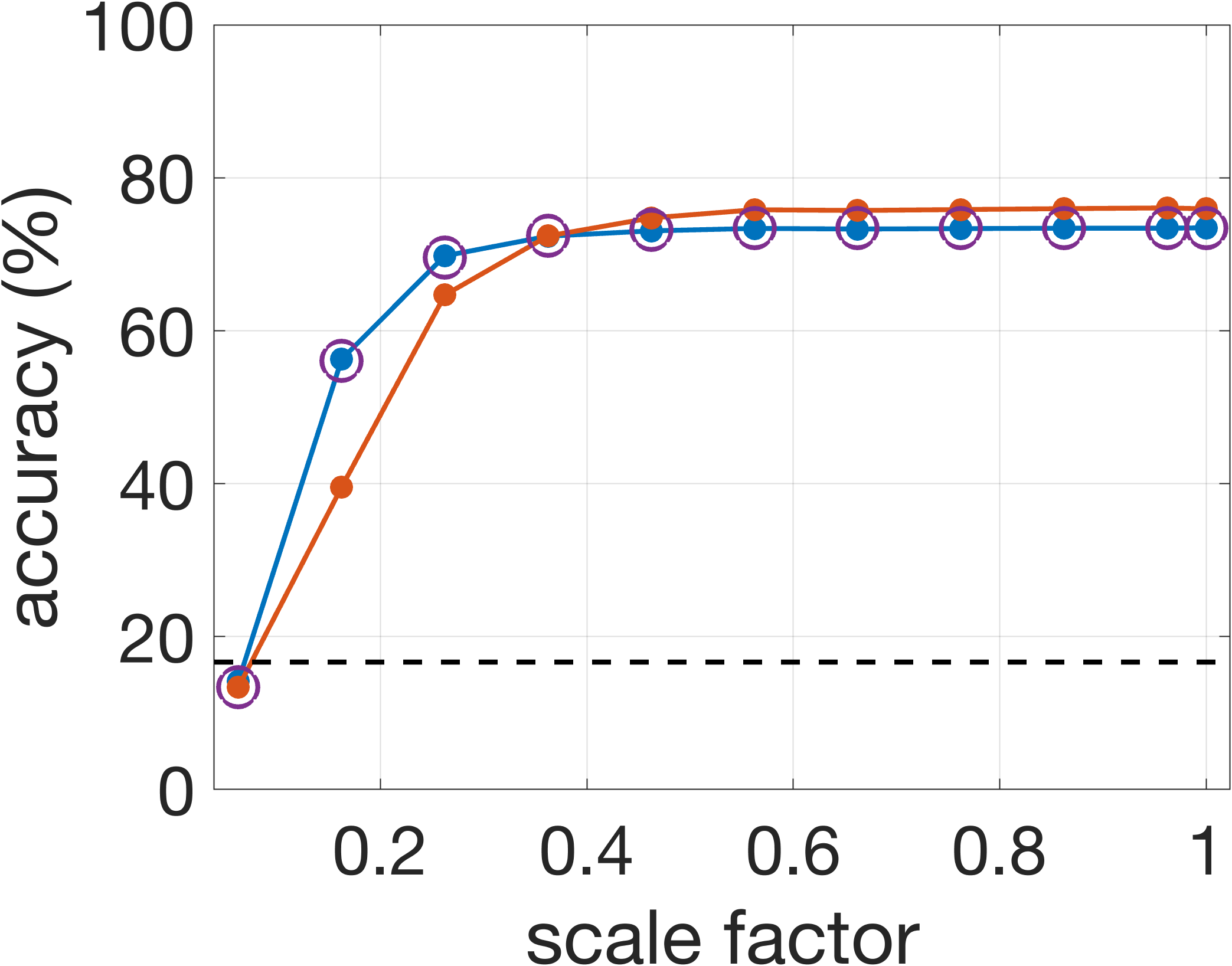} &
    \includegraphics[width=3.85cm]{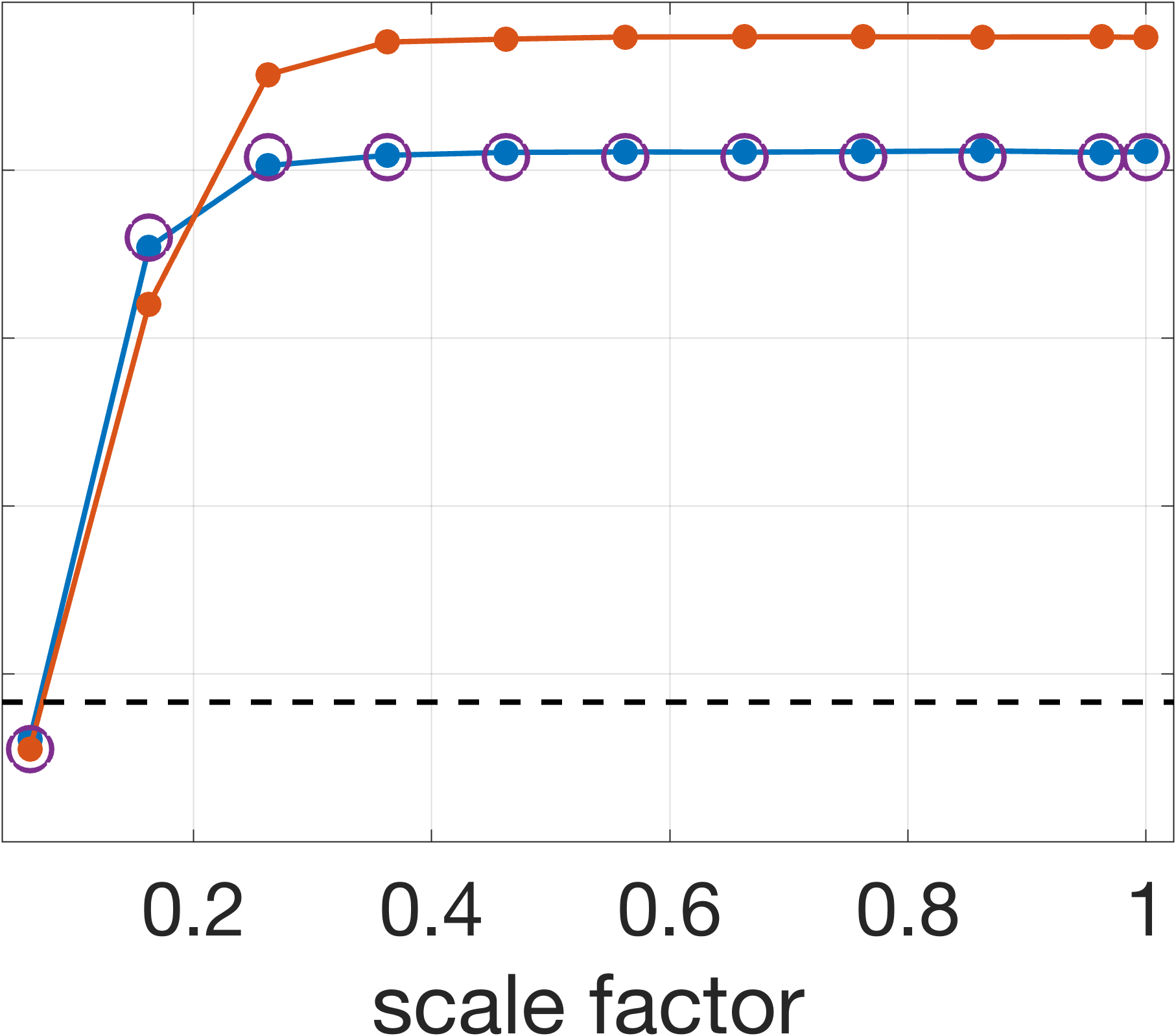} \\
    \includegraphics[width=4.4cm]{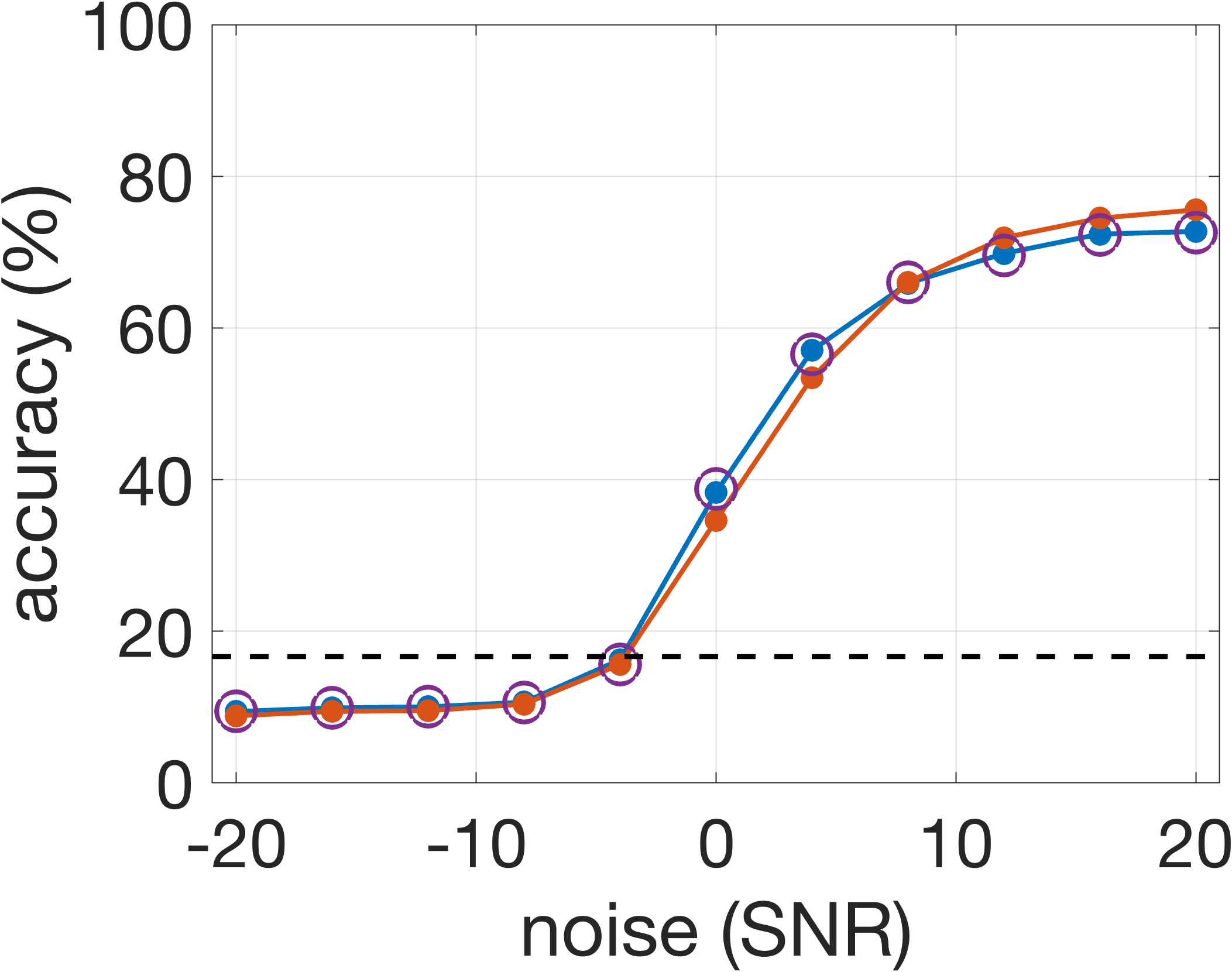} &
    \includegraphics[width=3.9cm]{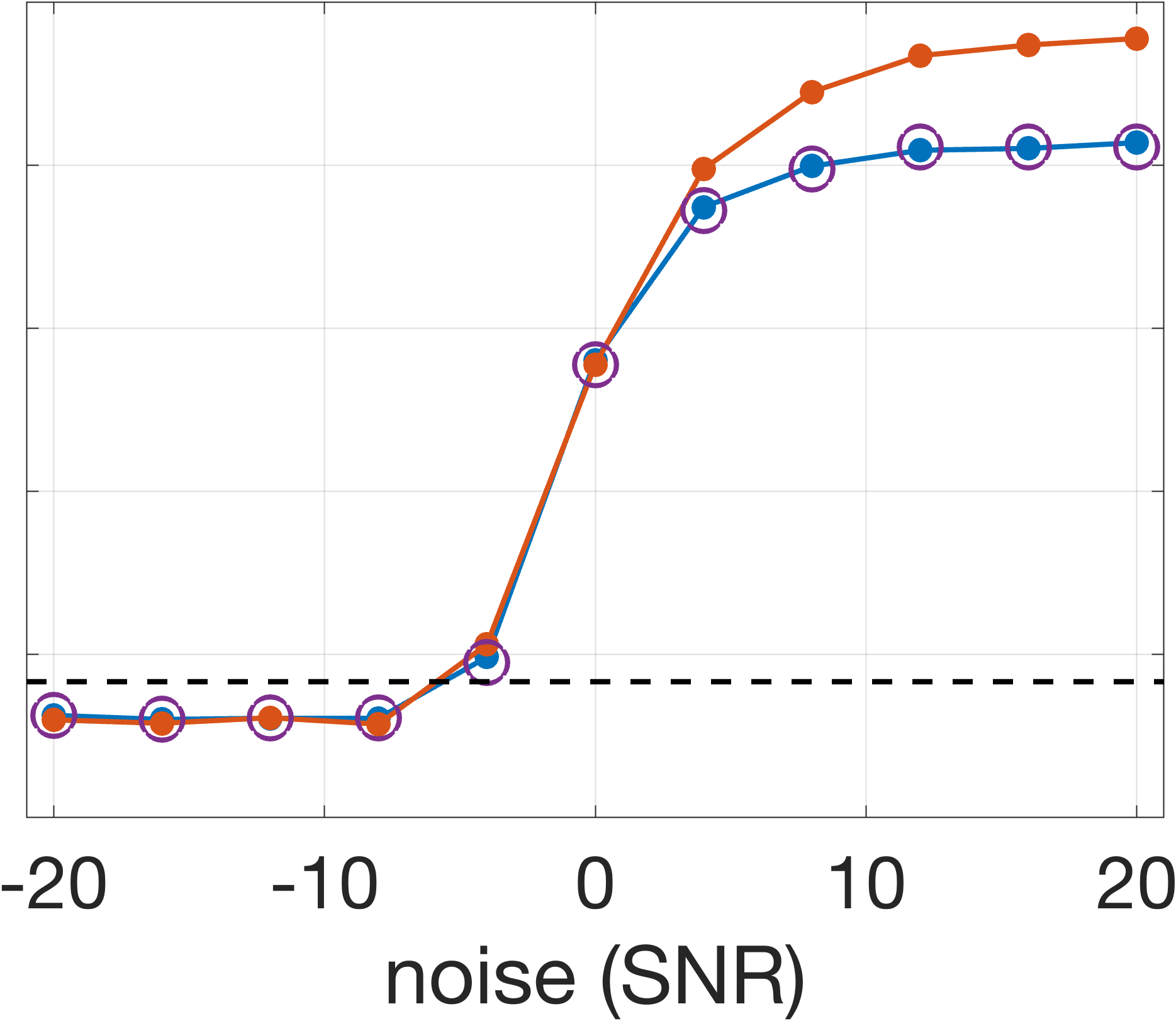} \\
    \includegraphics[width=4.4cm]{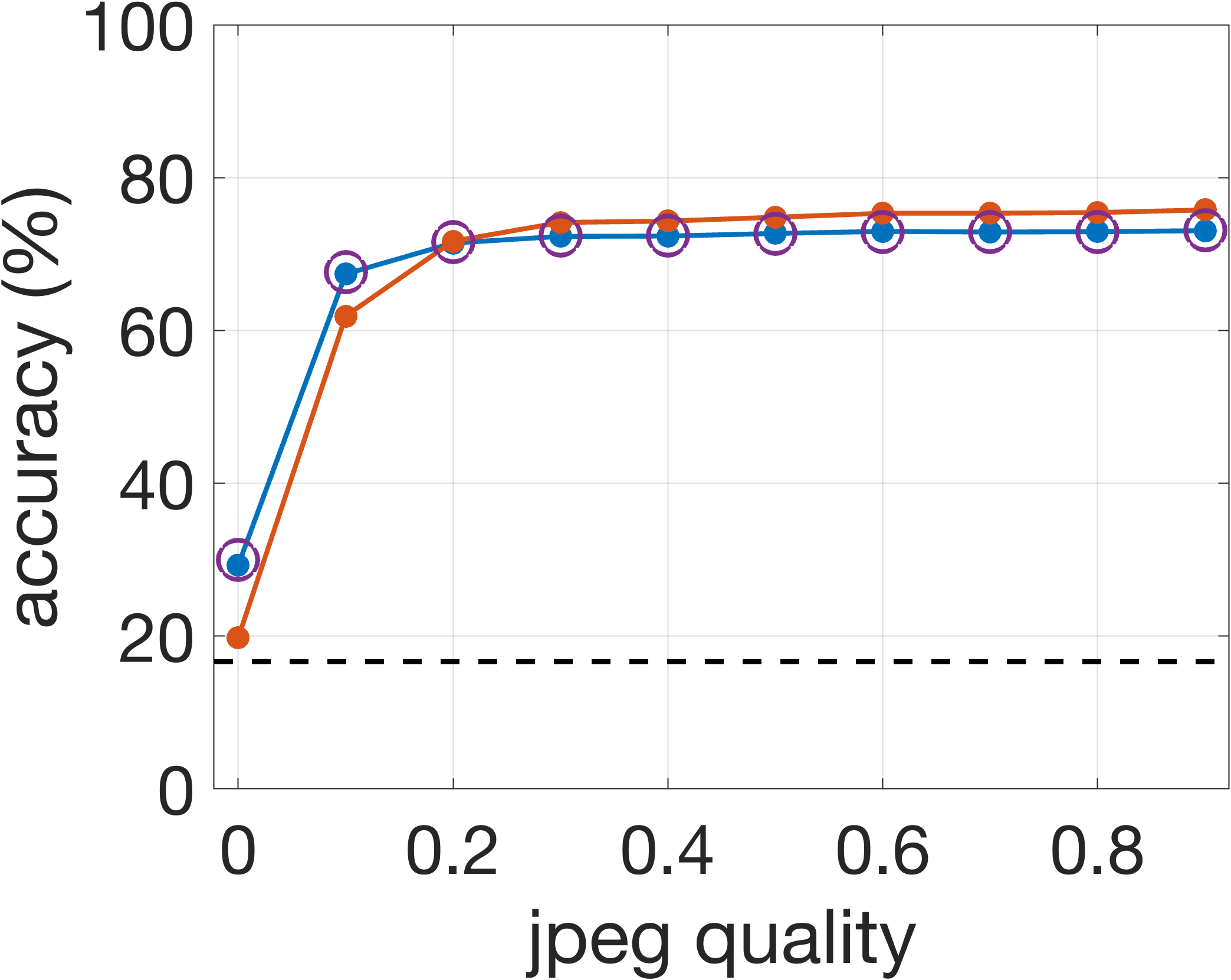 } &
    \includegraphics[width=3.82cm]{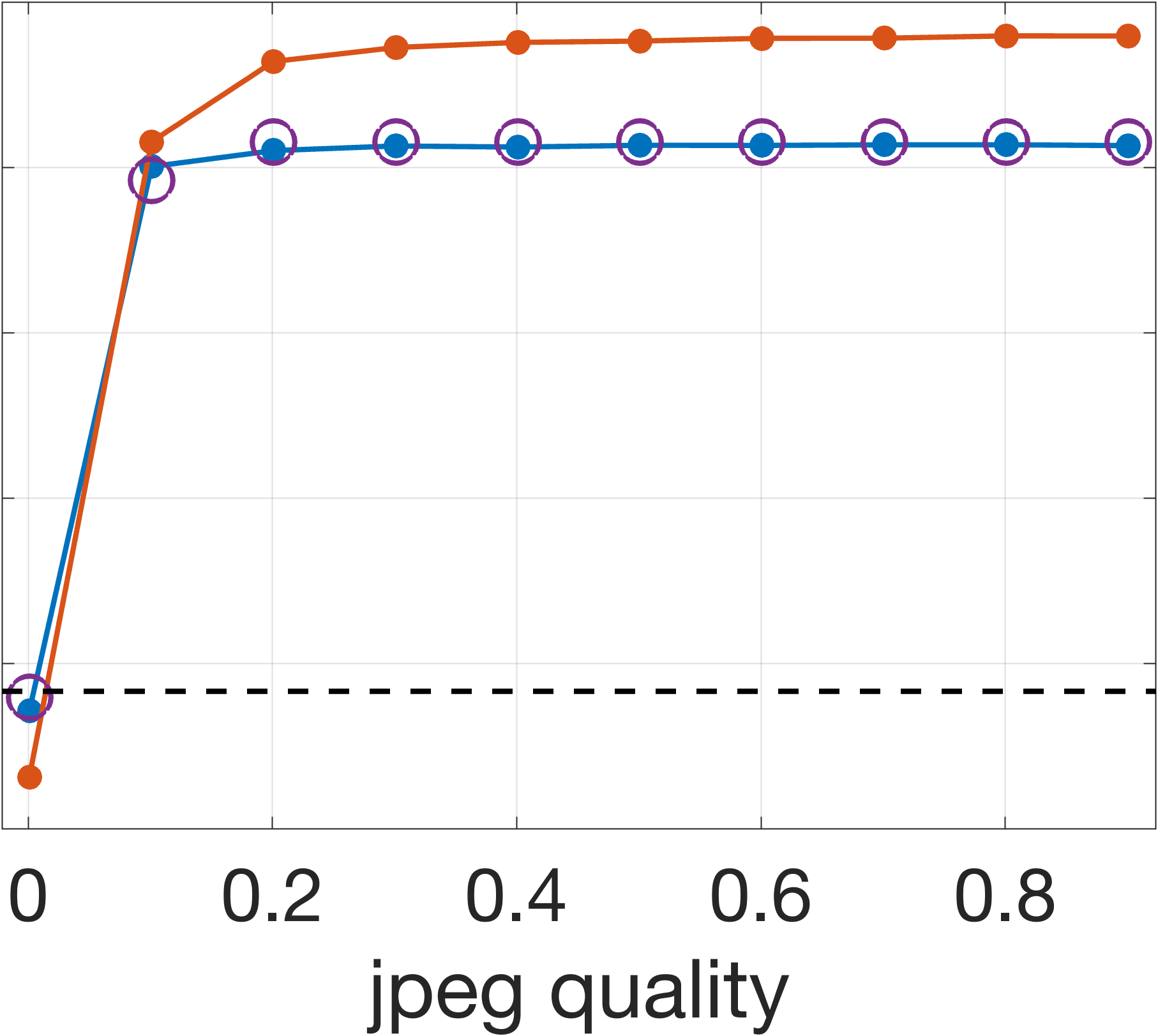 } \\
    \includegraphics[width=4.4cm]{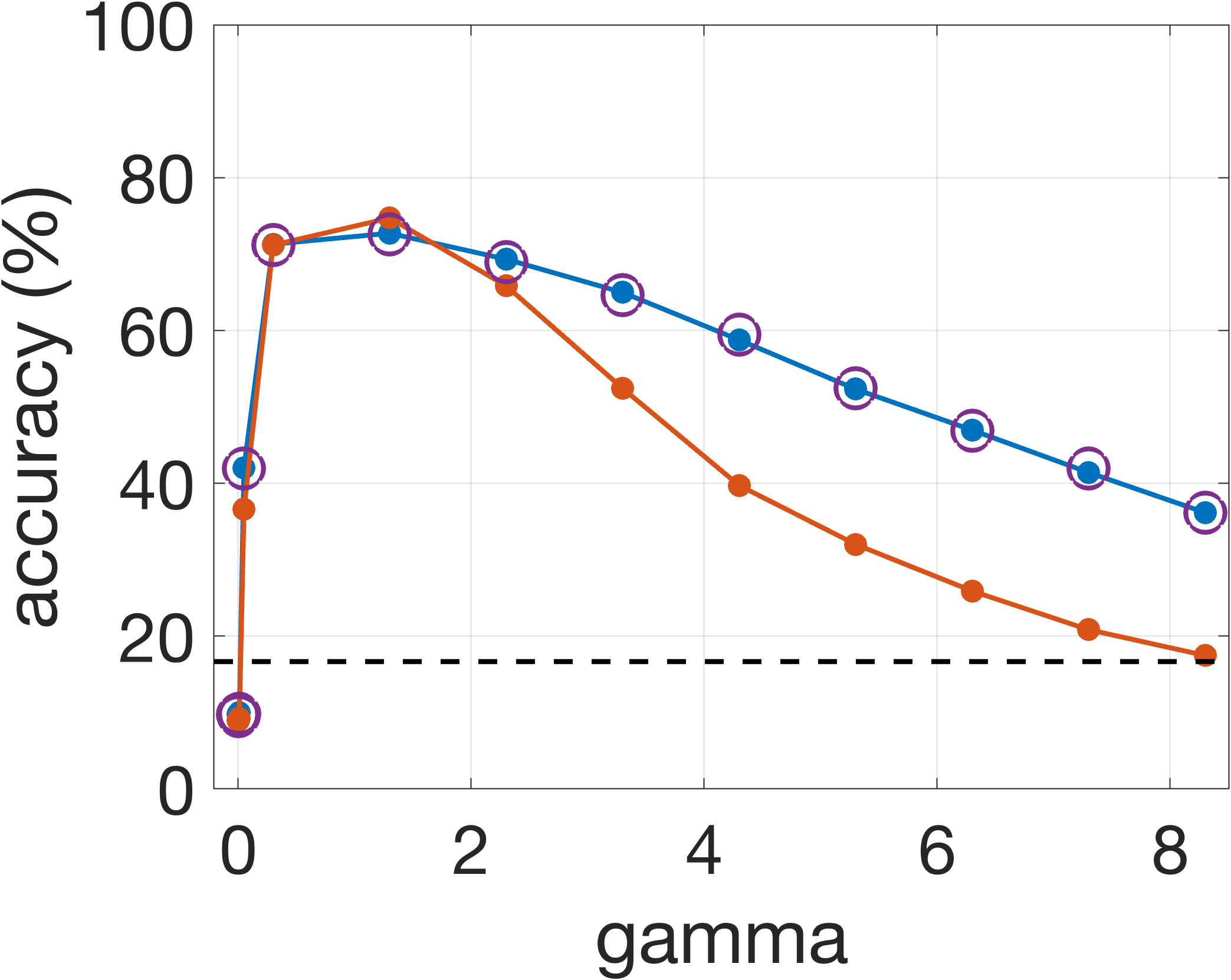} &
    \includegraphics[width=3.82cm]{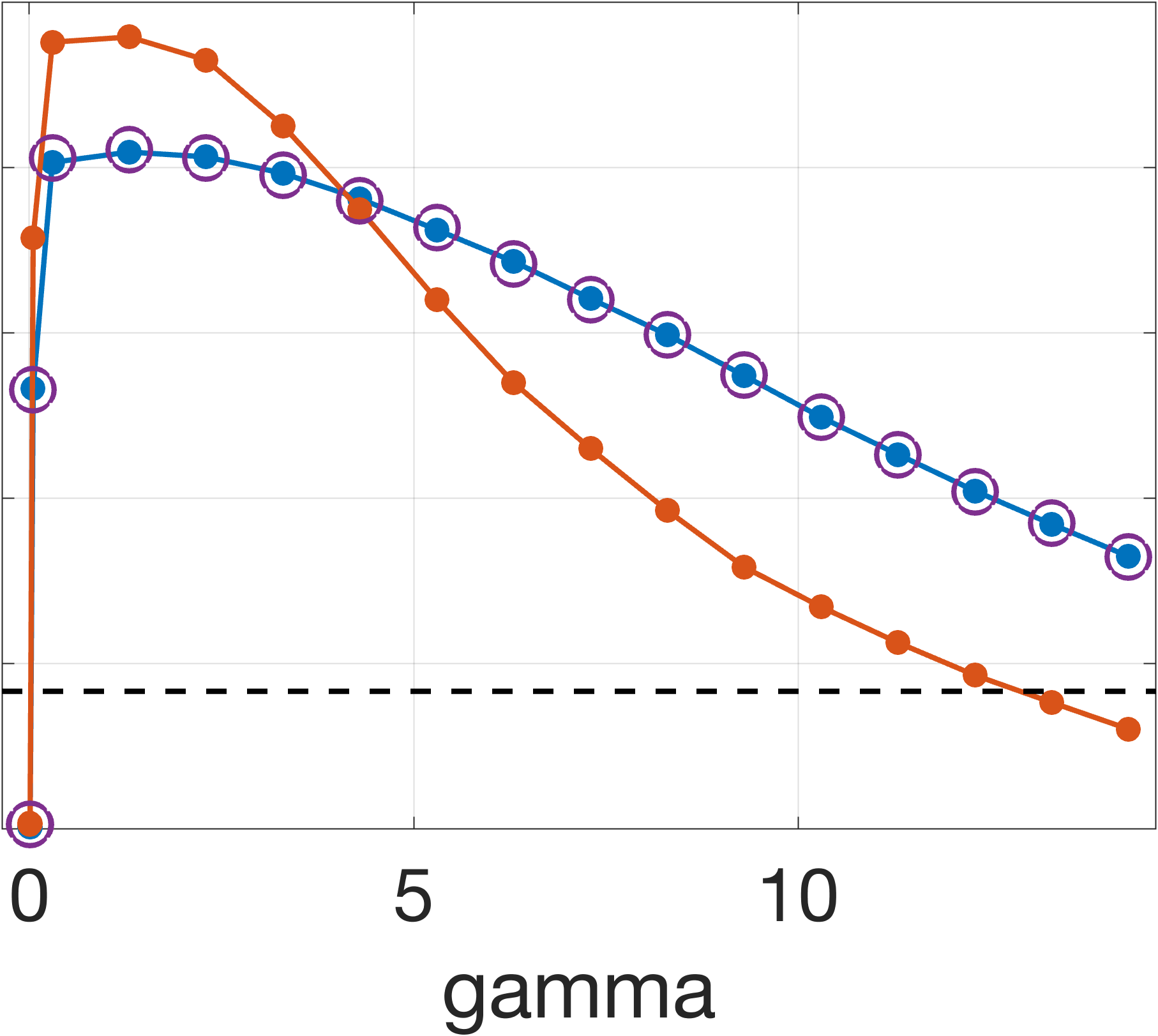} 
    \end{tabular}
    \caption{Accuracy on a forensic lineup task for FaceNet and ArcFace. Shown in each panel is the accuracy as a function of different degradations of the probe image. The filled blue circles correspond to the real-world CASIA-Webface dataset, the filled red circles correspond to the synthetic dataset, and the open magenta circles correspond to the calibrated synthetic dataset. Chance performance (horizontal dashed line) is $16.7\%$.}
    \label{fig:results-facenet-arcface}
\end{figure}

Shown in Fig.~\ref{fig:results-facenet-arcface} (blue) is the facial recognition accuracy on the real-world dataset as a function of image-based degradation of the probe image (resolution, blur, jpeg compression, noise, and gamma correction, Fig.~\ref{fig:degradations}). As expected the general trend is that accuracy  degrades as the probe image degrades in quality. Shown in Fig.~\ref{fig:results-facenet-arcface} (red) is the facial recognition accuracy on the synthetic dataset as a function of the same image-based degradations. 

We see that accuracy on the synthetic dataset can deviate substantially from the real-world dataset. These differences, however, can be calibrated for so that the synthetic dataset can be used as a reasonable proxy for the real-world dataset. This is done by adjusting the degradation parameters through a calibrated parametric lookup table that adjusts each parameter so that the accuracy on each level of degradation is matched between the synthetic and real-world datasets. Shown in Fig.~\ref{fig:results-facenet-arcface} (purple open symbols) is the result of this process in which we now see that the two datasets are well matched in terms of overall accuracy.

\begin{figure}[t!]
    \begin{center}
    \begin{tabular}{c}
        \includegraphics[width=0.95\linewidth]{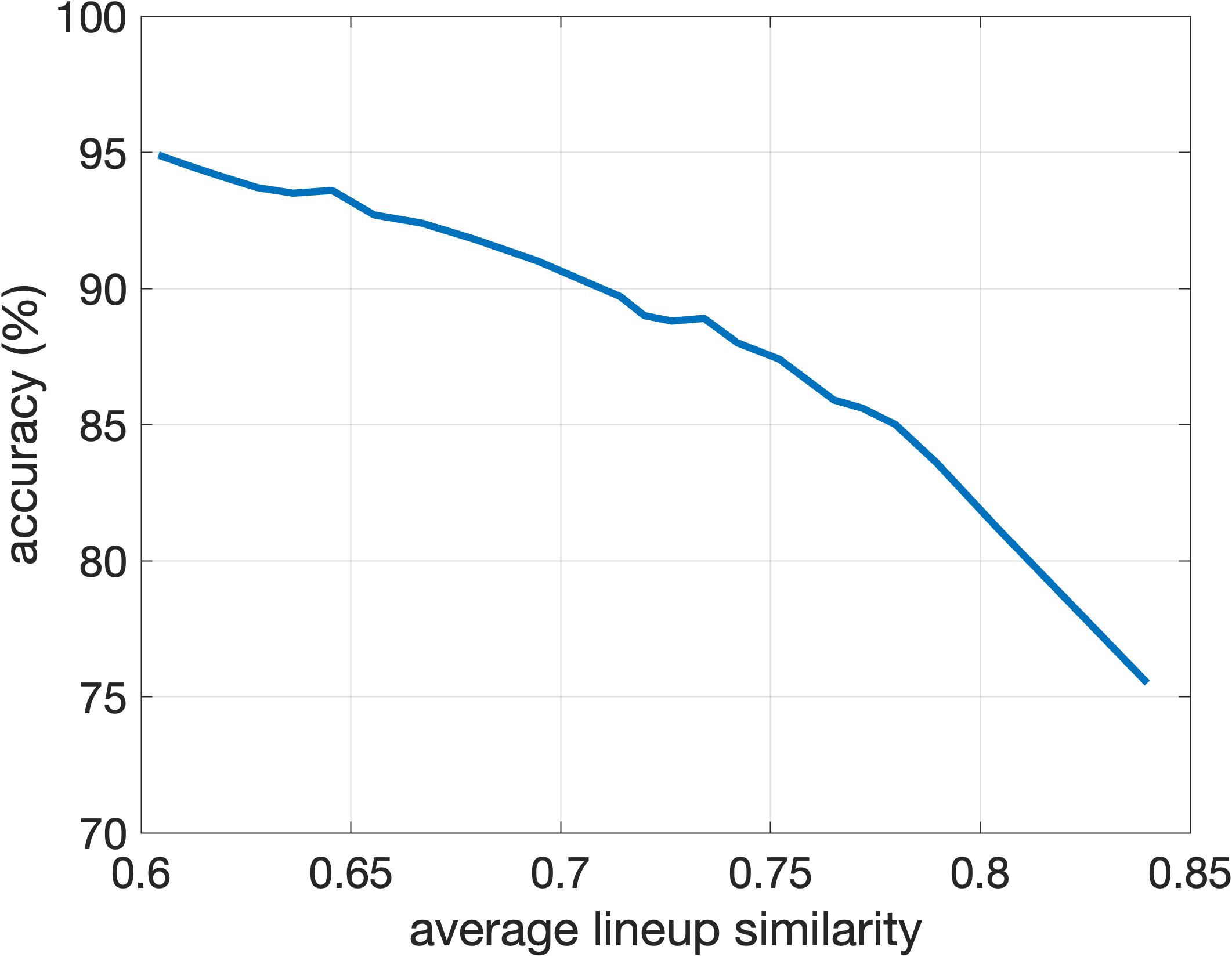}
    \end{tabular}
    \end{center}
    \vspace{-0.3cm}
    \caption{Forensic facial accuracy for FaceNet as a function of the average facial similarity between the probe and decoy faces. Facial similarity is measured as the cosine distance in the range $[-1,1]$ where lower values correspond to lower facial similarity. Accuracy improves from $75.5\%$ with the most similar decoys to $94.9\%$ for less similar decoys.}
    \label{fig:lineup}
\end{figure}

\subsection{Race and Gender}

We next evaluate race and gender differences across both network architectures for our synthetic dataset. Across race and (binary) gender, we see some discrepancies in overall accuracy for FaceNet. Base accuracy (i.e.,~no image degradation) for white people (n=$1659$) is $81.0\%$, while accuracy for black people (n=$208$) is $76.0\%$ and accuracy for all other ethnicities (n=$5633$) is  $74.1\%$. Accuracy for males (across all races) is $75.6\%$ and $75.9\%$ for females. We find no gender differences, but do find--consistent with previous studies--a small but not insignificant race bias in which white people are recognized at a higher accuracy.

These differences, however, largely vanish for ArcFace:  base accuracy for white people is $96.6\%$, while accuracy for black people is $96.2\%$ and accuracy for all other ethnicities people is $95.5\%$. Accuracy for males (across all races) is $96.1\%$ and $95.1\%$ for females.

\subsection{Lineup Similarity}

In the above results, the synthetic dataset forensic lineup was selected using the top five most similar faces, making the recognition task maximally difficult. Shown in Fig.~\ref{fig:lineup} is the impact of selecting a forensic lineup with a range of facial lineup similarities for the FaceNet architecture. This was done by sequentially shifting the five decoy faces in the lineup from the five most similar to increasingly less similar faces. Starting at a base accuracy of $75.5\%$, the accuracy rises to $94.9\%$ when the decoy faces are less similar to the identity being matched.

Although it is not surprising that identification accuracy increases as the decoys deviate in likeness relative to the probe, this result--with a $20$ percentage point shift--shows the importance of the lineup similarity in assessing the reliability of forensic facial recognition.

\subsection{Scene Degradations}

In the above, the synthetic dataset consists of faces with no occluding sunglasses or masks, and with relatively forward looking head poses. One of the benefits of the synthetic dataset is that we can--in a controlled setting--introduce facial variations to evaluate the robustness of these systems to expected real-world conditions.

For FaceNet, the baseline accuracy of $73.1\%$ dips to $68.2\%$ when the probe image contains opaque glasses and/or masks; dips to $70.5\%$ when the probe image contains more significant head rotations (greater than $30$ degrees in any of three orientations); and remains steady at $75.1\%$ when the probe image is taken in low light. In the presence of all three of these factors, accuracy drops to $63.1\%$.

ArcFace is more resilient to these degradations. The baseline accuracy of $95.6\%$  dips slightly to $91.4\%$ when the probe image contains dark glasses and/or masks; dips slightly to $93.8\%$ when the probe image contains more significant head rotations; and remains steady at $95.9\%$ when the probe image is taken in low light. In the presence of all three of these factors, accuracy dips to $88.4\%$.

%-------------------------------------------------------------------------
\section{Discussion}
\label{sec:discussion}

Facial recognition can play a critical role in law enforcement, but also raises thorny privacy, fairness, and bias issues. We regularly see academic and private-sector claims of highly accurate facial recognition systems which, in some cases, surpass human performance. We contend that these assessments may not accurately capture a real-world forensic scenario.

We contend that a large-scale synthetic facial dataset -- that affords careful control of the appearance and quality of facial images -- and a controlled forensic lineup -- that affords careful control of the comparison group -- affords a more accurate accounting of the accuracy of facial recognition. Previously reported accuracies for FaceNet and ArcFace ranging between $95\%$ to $99\%$ drop by more than $20$ and $10$ percentage points to $73\%$ and $83\%$. More challenging probe faces, with large head rotations and occluding glasses and masks, leads to a further decrease in accuracy for FaceNet to $63\%$.

In the forensic scenario where a suspect is being matched to an in-the-wild photo, this drop in accuracy is significant and should give some pause as to the reliability of these facial recognition systems in challenging more forensic identification tasks.

While synthetic datasets provide a useful mechanism to carefully examine the accuracy of facial recognition, we have shown that accuracy can vary greatly from real-world datasets. The type of calibration performed in Fig.~\ref{fig:results-facenet-arcface}, however, can adjust for these differences. 

While the past three decades have seen impressive improvements in facial recognition, great care should be taken in the high-stakes deployment of these technologies. Our analysis has shown that the accuracy on challenging forensic facial recognition can significantly vary from reported accuracies.

%-------------------------------------------------------------------------

\bibliographystyle{ieee_fullname}
\bibliography{main}

\end{document}